\documentclass[letterpaper, inpress]{jds} 

\jdsmonth{May}                 
\jdsyear{2022}                  

\jdsreceived{May, 2022}       
\jdsaccepted{}         

\usepackage{amsfonts,amsmath,amssymb,amsthm}
\usepackage{booktabs}

\usepackage{bbm}
\usepackage{dsfont}
\usepackage{mathrsfs}
\usepackage[english]{babel}
\usepackage[utf8]{inputenc}
\usepackage{subcaption}
\usepackage{bm}

\usepackage[utf8]{inputenc}
\usepackage{booktabs,lipsum}

\newcommand{\vx}{\mathbf{x}}

\newcommand{\vy}{\mathbf{y}}
\newcommand{\vY}{\mathbf{Y}}

\newcommand{\nmathbf}{\bm}

\newcommand{\sD}{\mathscr{D}_n}

\newcommand{\sH}{\mathscr{H}}

\newcommand{\sX}{\mathscr{X}}
\newcommand{\sY}{\mathscr{Y}}

\newcommand{\loss}{\mathcal{L}}

\def\bf0{\nmathbf 0}

\def\bfbeta{\nmathbf \beta}

\usepackage{lipsum}

\title[kNNvs]{ Improving the Predictive Performances of  $k$ Nearest Neighbors Learning by Efficient Variable Selection}

\author[1]{Eddie Pei\thanks{Corresponding author. Email: ep2667@rit.edu}}
\author[2]{Ernest Fokou\'e }
\affil[1]{Munsell Color Science Laboratory, Rochester Institute of Technology, 85 Lomb Memorial Drive, Rochester, New York 14623}
\affil[2]{School of Mathematical Sciences, Rochester Institute of Technology, 85 Lomb Memorial Drive, Rochester, New York 14623}

\begin{document}

\maketitle

\begin{abstract}
  This paper computationally demonstrates a sharp improvement in predictive performance for $k$ nearest neighbors thanks to an efficient forward selection of the predictor variables. We show both simulated and real-world data that this novel repeatedly approaches outperformance regression models under stepwise selection.\@
\end{abstract}

\begin{keywords} 
  $k$ Nearest Neighbors;
  Variable selection;
  Classification;
  Regression;
  Grid Search.
\end{keywords}

\section{Introduction}
\label{sec:intro}

Variable selection methods play a critical role in machine learning, also known as feature selection. Variable selection helps avoid the curse of dimensionality, improving the prediction performance, shorter the training time or saving the computer resources, and so on. $k$ nearest neighbors (kNN) is a non-parametric classification method introduced by Evelyn Fix and Joseph Hodges in 1951 \citep{kNN}. Due to its simplicity and flexibility, it is a viral algorithm. The applications of the kNN algorithm includes lots of areas, for instance visual recognition \citep{Application:visual}, text categorization \citep{Application:text} \citep{Application:text2} MRI brain cancer classification \citep{Application:brain} and so on. It is indeed a powerful machine.
$k$ nearest neighbors is a very well-studied algorithm. There are numerous researches about $k$ nearest neighbors. When people use kNN, everyone would consider an optimum $k$ but barely consider the best subset variables in kNN. So here, we propose a grid search variable selection of kNN along with an R package. Using grid search among the input function space, carefully choose the loss function and compare the loss to choose the best subset variables.

\section{$k$ Nearest Neighbor with Grid Search Variable Selection }

\label{sec:$k$ Nearest Neighbor with Grid Search Variable Selection}

For a machine learning task, one trying to build a function $f$ : $\sX \rightarrow \sY$ mapping input space $\sX$ to the output space $\sY$ \citep{FokoueErnest:2020:1}. A theoretical risk function defined as 
$$
    R(f)=\mathbb{E}\Big[\loss(\vy,f(\vx)\Big]= \int_{\sX*\sY}\loss(\vx,\vy)dp(\vx,\vy)
$$

where $\loss(\cdot,\cdot)$ is the loss function, ideally one wants to achieve the goal to find the universal best function 
$$f^* = \underset{\sY^\sX}{\tt arginf}\Big{\{}\mathbb{E}[\loss(y,f(\vx)]\Big{\}}= \underset{\sY^\sX}{\tt arginf}\Big{\{}\int_{\sX*\sY}\loss(\vx,\vy)dp(\vx,\vy) \Big{\}}$$

but in practically speaking given a data set $\{(\vx_1,\vy_1),(\vx_2,\vy_2)...(\vx_n,\vy_n)\}$, it is impossible for one to find the joint distribution $P(\vx,\vy)$,so instead of trying to search the function space $\sH$ to find $f^*$, in reality one could define a empirical risk function $\widehat{R}(f)$ as following
$$\widehat{R}_n (f)= \frac{1}{n} \sum_{i=1}^n \loss\Big(y_i,f(\vx_i)\Big)$$ 
and then find the best empirical function $\widehat{f_n}$ within the function space $\sH$, namely 
$$\widehat{f_n}=\underset{f \in \sH}{\tt arginf}\Big\{\frac{1}{n} \sum_{i=1}^n \loss(y_i,f(\vx_i)\Big\}$$

In this case, the function space $\sH$ contains all the kNN models for one data set, denoted as $\sH_{kNN}$; based on the loss, we pick the best model among all. We are using Grid Search to scan the input space to find the different combinations of variables to achieve the variable selection. Speak of the Grid search, which is usually used as a tuning technique that attempts to find the optimum values of hyperparameters. It is simply an exhaustive search through a manually specified subset of the hyperparameter. In this paper, we would use the "grid search" idea to search through the input space to find the best subsets. Grid search suffers from the curse of dimensionality, which sounds like a very computationally expensive technique, but kNN has an excellent reputation known for its simplicity which helps to solve the dimensionality problem. To test on this problem, we create a particular case to test the dimensionality vs. running time in next section.

Speaking of $k$ nearest neighbors algorithm, it could be described as the following steps: first, select the number $k$ of the neighbors; then calculate the distance, take the $k$ nearest neighbors, next Among these $k$ neighbors, count the number of the data points in each category; last assign the new data points to that category for which the number of the neighbor is maximum. Shown in algorithm \ref{KNN-alg}.

\begin{algorithm}
\kwInput { $k$, \text{the number of neighbors}; \text{$\sD$ is the data set}; $\vx_{new}$, \text{the new data point}  }
\kwOutput{the label of $\vx_{new}$}
\For{$i \in (1 : n)$}{
\sf $d(\vx\textsubscript{new}, \vx\textsubscript{i}) \gets compute$
           }
Take the k nearest neighbors;
\\
$\widehat{Y_{new}}$ = argmax\Big($Y_i$, i $\in$
\text{\{$k$ nearest neighbors\}\Big)}
\caption{$k$ Nearest Neighbor}
\label{KNN-alg}
\end{algorithm}

Indeed given classification data $\sD= \{(\vx_i,y_i) \overset{iid}{\sim} p_{xy}(\vx,y),\,\, \vx_i \in \sX,\, y_i\in \{1,2,\cdots g \cdots G\}, \,\, i=1,\cdots,n\}$, and a new data point $(\vx_{new},y_{new})$, one wants to predict the label for new data point $\widehat{y_{new}}$ we shall herein express $k$ nearest neighbors learning models in the following form 
\begin{equation}
   \widehat{f}_n^{kNN} (\vx_{new}) = \underset{{g=1,2 \cdots G}_{t}}{{\tt argmax}} \{\widehat{\pi_g^{(k)}(\vx_{new})} \}
   \label{eq:kNN}
\end{equation}

where k is the number of neighbors and  $\widehat{\pi_g^{(k)}(\vx_{new})}$ could be written as 
\begin{equation}
 \widehat{\pi_g^{(k)}(\vx_{new})} = \frac{1}{k} \sum_{i=1}^n\mathbb{1}(\vx_i \in V_k(\vx)) \mathbb{1}(y_i = g)   
\end{equation}
To identify the neighbors which is based on the the distance, according to the type of prediction problem, one could wisely choose a suitable distance. Euclidean distance \citep{Euclideandistance}; also know as the $l^2$ distance simply define by $d(\vx_i,\vx_j)=\sqrt{\sum_{l=1}^q(\vx_{il}-\vx_{ijl})^2}=||\vx_i-\vx_j||_2$, it is one of the very commonly used distance.Like we said earlier, there are many kinds of distance, in here just to mention a few as following: manhattan distance \citep{manhattandistance}, also known as $l_1$ distance, which is written as  $d(\vx_i,\vx_j)=\sum_{l=1}^q|\vx_{il}-\vx_{ijl}|=||\vx_i-\vx_j||_1$; Minkowski distance $d(\vx_i,\vx_j)=\Big\{\sum_{l=1}^q|\vx_{il}-\vx_{jl}|^p\Big\}^{1/p}$, it is known as $l_p$ distance; Jaccard/Tanimoto distance \citep{Jaccarddistance} $d(\vx_i,\vx_j)=1-\frac{\vx_{i} \times \vx_{i}}{|\vx_{i}|^2-|\vx_{i}|^2-\vx_{i}\times\vx_{j}}$ where $\vx$ is the binary vector ie $\vx_i \in \{0,1\}^q$. Choose a suitable distance could benefit the kNN performance.  

Variable selection can be treated as a technique that an evaluation measure that scores the different feature subsets in order to search for the new variable subsets to achieve the machine learning goal, for instance, improvement of the prediction performances. In this paper, we propose a grid search variable selection for $k$ nearest neighbors to improve prediction performance.  

$k$ Nearest Neighbor with grid search variables selection could be defined as the following steps: first, for the preparation, same as $k$ nearest neighbors, we need to pick $k$, and also we need to divide the datasets into training and test sets; next create the grid, based on the input space, for instance, the dimension of the input space is $p$, so we create $p$ level kNN models, for each level of the models we compare the loss, pick the best model, then compare the number of $p$ model's loss, pick the best one which is the final model with the best subset variables. The loss includes two kinds, for regression, the loss is the Mean Squared Error; for the classification, the loss we used here is accuracy.

There is a pronounced problem in the algorithm, which is in step to create $p$ levels kNN models. If we use the traditional grid search, for the number of $p$ input space, we need to create the number of $C_p^1+C_p^2+\cdots+C_p^p$ models, which is extremely computationally expensive. To reduce the cost here, inspired by the forwarding selection technique, instead of creating all the grids at the same time, we start from one variable model, pick the best one-variable model then, based on the best one-variable model, build the two-variable model, and repeat this process until the full model. In this way, we could reduce the number of models to $p+(p-1)+\cdots+1$, which is a tremendous reduction in the number of models. One might still say there are many models to be built, but we should keep this in mind: "kNN is known for its simplicity." That is why it can be achieved better on "kNN" instead of some very complex machine such as Neural Networks. In the next section, we will provide a simulation study over a comparison for the dimensionality vs. running times.

\begin{algorithm}
\kwInput{ \begin{itemize}
    \item $\sD1$ and $\sD2$, Training data and Test data;
    \item $k$, the number of neighbors;
    \item {$\vx_{input}$=$\phi$}, \text{the model start with the empty  model};
    \item $B=\{1:p$\}, $B$ as bag is a bridge variable using to select the variables. 
\end{itemize}}
\For{i in 1 : p}
    {
    \For{j in B}{
         $\vx_{input} = \{\vx_{input},\vx_j\}$ \\
         $\text{run kNN, calculate} \ \loss(.)_j$ \\
         $\loss(.)_{i} = Optimal\Big(\loss(.)_j, j \in \{1: length(B_i)\}\Big) $ \\
         $\vx_{i} \ \text{is the} \ \vx_j \ \text{pair with} \  pcc_{i} $\\
                }
      $\vx_{input} = \{\vx_{input},\vx_{wk}\}$ ; \ $\loss(.) = \{\loss(.), \loss(.)_{wi} \}$ ; $list_i =  \vx_{input}$ \\     
                }
\kwOutput{
\begin{itemize}
    \item $\loss(.)_{w} = Optimal \Big ( \loss(.)_i, i \in \{1:p\} \Big )$
    \item Selected Variable = $list_w$, which is the $list_w$ pair with  $\loss(.)_w$
\end{itemize}
}
\caption{$k$ Nearest Neighbor with grid search variable selection}
\label{alg:kNNvs}
\end{algorithm}

\section{$k$ Nearest Neighbor with Grid Search Variable Selection R package}

$k$ Nearest Neighbor with Grid Search Variable Selection package is called "kNNvs", which is an R package \citep{kNNvs:R:package}, try to solve the variable selection for kNN classification or kNN regression in order to improve the predictive performance and identifying the important features. It follows the ideas of algorithm \ref{alg:kNNvs}. In this package there are 6 input features, which are "train\_x", "test\_x", "cl\_train","cl\_test", "$k$" and "model":
\begin{itemize}
    \item train\_x : the training predictor variables, matrix or data frame; 
    \item test\_x : the test predictor variables, matrix or data frame; 
    \item cl\_train : the training response variable;
    \item cl\_test : the test response variable;
    \item $k$ : is the number of neighbors;
    \item model: it includes regression or classification.
\end{itemize}
The "cl\_test" is an optional feature here. If included "cl\_test," then the function will use the training data as training and test data to find the best combination of the variable along with the best loss. However, if it does not include "cl\_test," the function will divide the training data into 70\% as a new training set and 30\% as a test set to make the variable selection and apply the new model to the "test\_x" to get the estimated value $yhat$. For the output value, in the first case, it includes each level best variable combination along with the loss, best variable combination, the best loss, which is based on the testing data, and the estimated value $yhat$; for the second case, since there are no test response variables, so the best acc or MSE is the training best acc or MSE. Also, in this package, we provided two examples using the public data "iris3" to demonstrate these two cases. See appendix. 

\section{Computational Explorations and Demonstrations}

In this section, we use simulated and real-world data to test the model's performance. For the simulation study, there are classification and regression cases. First, we create the simulation classification data with noise variables and compare $k$ nearest neighbors with grid search variable selection with stepwise logistic regression \citep{logisticregression} to see the prediction performance and the selected variables. Secondly, we test the simulation regression data with noise variables and compare $k$ nearest neighbors with grid search variable selection with stepwise multi-linear regression based on the prediction performance and the selected variables. Last part of the simulation study, we generate data with different sizes of input space to check the running time, including classification and regression cases. For the real-world data, there are three sections: first, we want to see the prediction performance of $k$ nearest neighbors with grid search variable selection, so we compare it with other powerful machines, such as $k$ nearest neighbors, support vector machine with Gaussian kernel \citep{Ring:2016:AGR}, support vector machine with linear kernel, support vector machine with polynomial kernel \citep{Wahba:1998} \citep{scholkopf2018learning} and Gaussian process machine \citep{GP:Book:1}; next, we compare the $k$ nearest neighbors with grid search variable selection with stepwise logistic regression to see the performance and the selected variables, the last experiment is comparing kNNvs with step multi-linear regression on the real-world regression data evaluated by the predictive performance and selected variables.

\subsection{simulation study}

\subsubsection{Classification Simulation Study}
In the classification simulation study, we generate 200 observations, all the obs are generated from a multi-normal distribution, there are 10 predictor variables $\{\vx_1,\vx_2 \cdots\vx_{10}\}$, and 5 of them ($\vx_1,\vx_2...\vx_5)$ are used to generate response variables $Y$, and 5 of them ($\vx_6,\vx_7...\vx_{10})$ are the noise variables. The response variable $Y$ is generated from a Bernoulli distribution is given by 
\begin{equation}
    (\vY|\vx)\sim {\sf Bernoulli}(\pi(\vx)) \quad  \text{where} \quad \pi(\vx)=\Pr(\vY=1|\vx)=\frac{1}{1+e^{-\vx^\top\bfbeta}}
    \label{eq:binarydata}
\end{equation}
where $\{\vx=(\vx_1,\vx_2,\vx_3,\vx_4,\vx_5), \bfbeta=(\bfbeta_1,\bfbeta_2,\bfbeta_3,\bfbeta_4,\bfbeta_5)^T\}$. Since we want to compare the variable selection methods, so we random the order of $\vx_1$ to $\vx_{10}$ 's order. We compare the step-logistic regression with KNNvs. For the data set, we divide the data into 70\% as training set and 30\% as test set, and run 50 times to compare the accuracy on the test set and selected variables  

\begin{figure}[!ht]
    \centering
        \begin{subfigure}[b]{0.32\textwidth}
        \includegraphics[width=\textwidth]{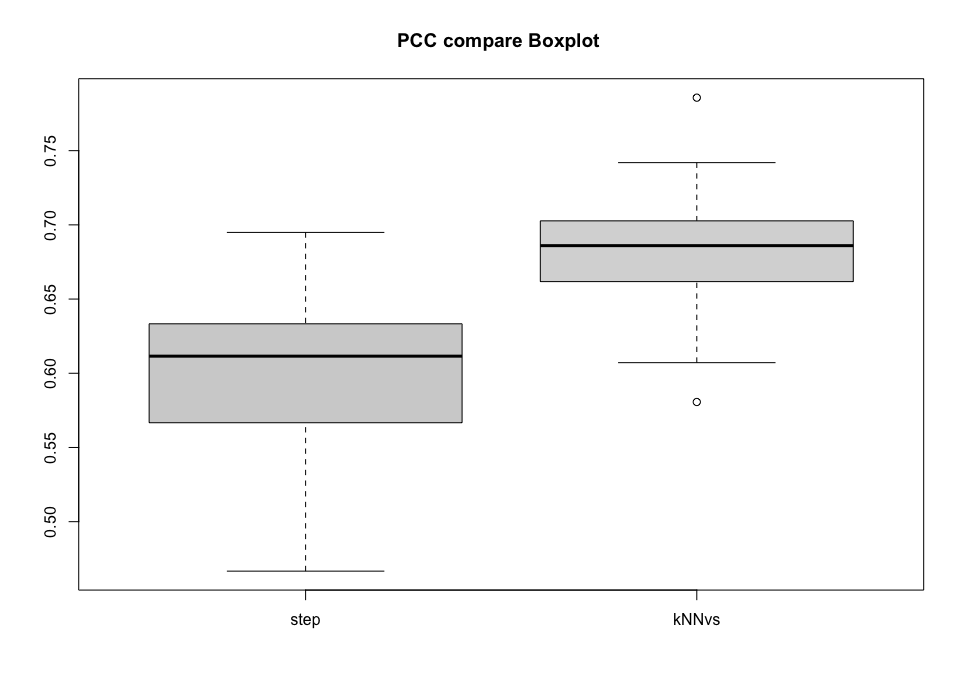}
        \caption{the boxplot }
        \label{fig:compare with slr sim}
    \end{subfigure}
        \begin{subfigure}[b]{0.33\textwidth}
        \includegraphics[width=\textwidth]{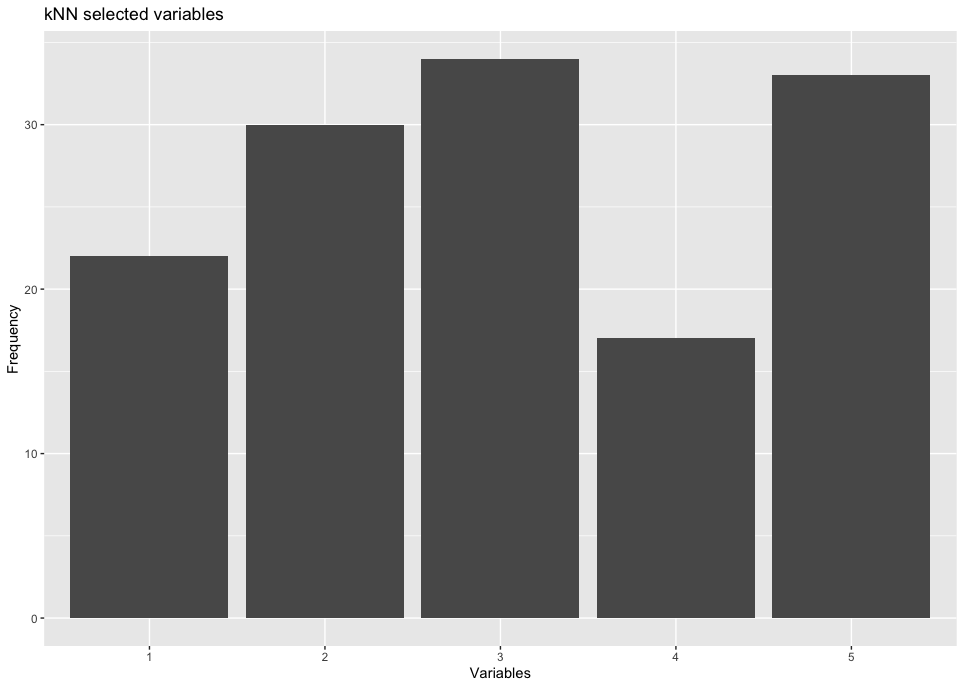}
        \caption{kNNvs }
        \label{fig:his_knnvs_sim}
    \end{subfigure}
    \begin{subfigure}[b]{0.33\textwidth}
        \includegraphics[width=\textwidth]{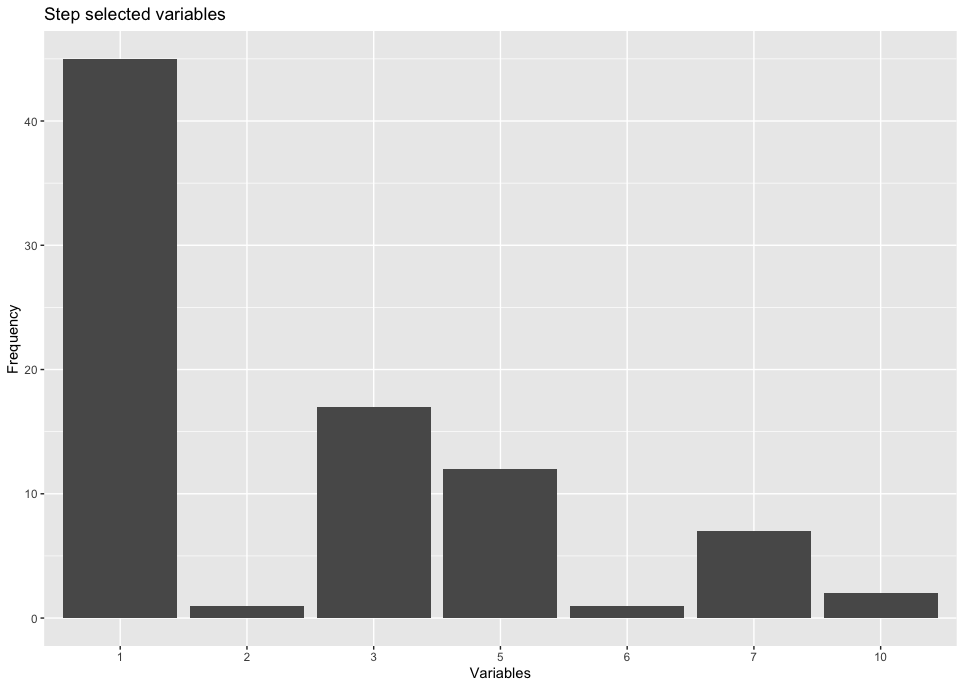}
        \caption{step}
        \label{fig:his_step_sim}
    \end{subfigure}
    \caption{comparison between kNNvs and stepwise logistic regression and the frequency of selected variables}
      \label{fig:his_com_sim}
\end{figure}

Figure \ref{fig:his_com_sim} shows the results. Figure \ref{fig:compare with slr sim} shows the accuracy of 50 runs. According to the figure, kNNvs has a smaller range, and the median is higher than step logistic regression. Furthermore, it shows that the kNNvs has a better prediction performance than step logistic regression. For the comparison of selected variables, Figure \ref{fig:his_com_sim} shows 50-runs of selected variables' frequency. Based on the figure, kNNvs could always select the right variable, it never selects the noise variable 50 times, but for the stepwise logistic regression, it catches up with noise variables $\vx_6, \vx_7$ and $\vx_{10}$. So based on these comparison results, we could conclude that kNNVs are better on prediction performance and variable selection. 

\subsection{Regression Simulation Study}
In the regression simulation study, we generate 100 observations, 9 predictor variables  $\{\vx_1,\vx_2,\vx_3,\vx_4,\vx_5,\vx_6,\vx_7,\vx_8,\vx_9\}$, among them $\{\vx_2,\vx_4,\vx_5,\vx_6,\vx_8\}$ are the noise variables, and the response variable is the combination of $\{\vx_1,\vx_3,\vx_7,\vx_9\}$ and noise$\epsilon$ which is generated from a normal distribution, the response variable $Y$ is given by:
\begin{equation*}
    \vY= 0.13\times \vx_1-0.5\times \vx_3-0.17\times\vx_7\times\vx_9 + \epsilon
\end{equation*}
in the regression case, we compare kNNvs with step regression, we randomly divided data into 70\% as training and 30\% as test data, and do 50 times replications. The model evaluation is based on the test error: mean squared error(MSE).

\begin{figure}[!ht]
    \centering
        \begin{subfigure}[b]{0.32\textwidth}
        \includegraphics[width=\textwidth]{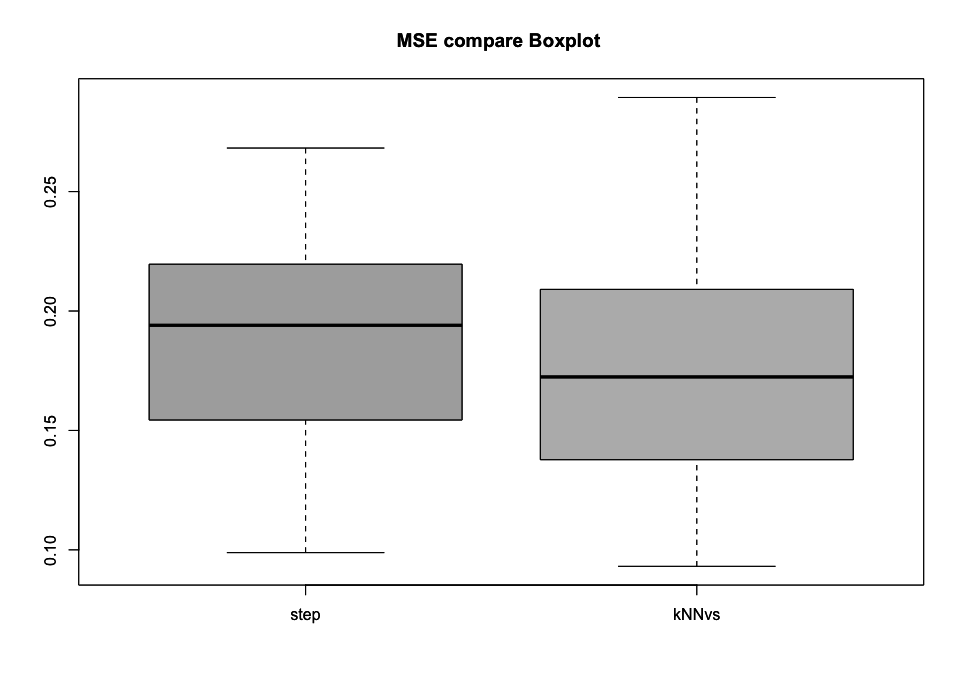}
        \caption{the boxplot }
        \label{fig:compare with step sim for regression}
    \end{subfigure}
        \begin{subfigure}[b]{0.33\textwidth}
        \includegraphics[width=\textwidth]{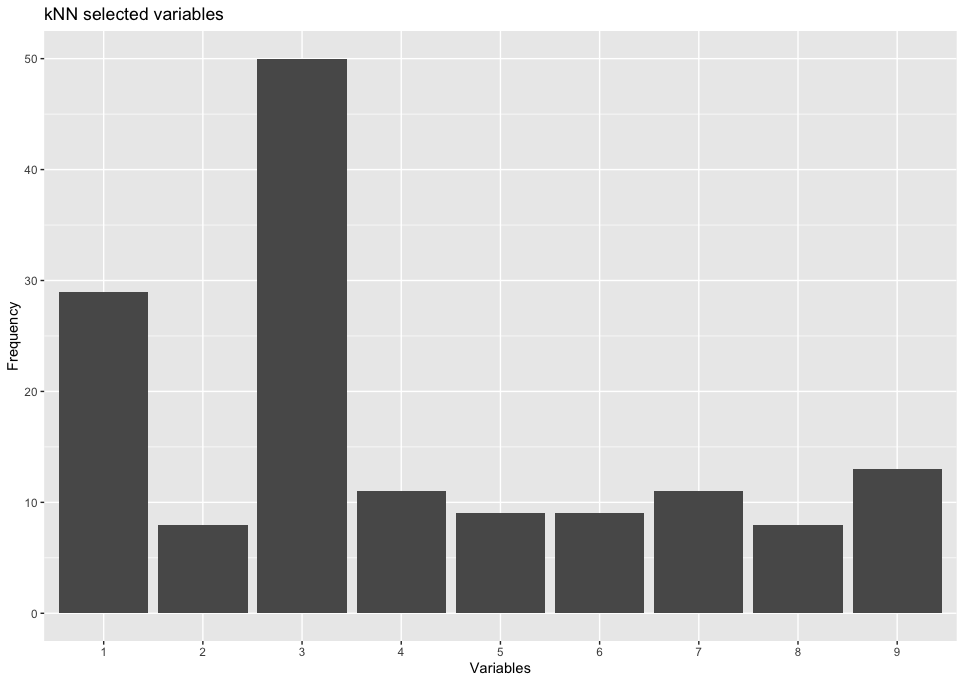}
        \caption{kNNvs }
        \label{fig:his_knnvs_sim-re}
    \end{subfigure}
    \begin{subfigure}[b]{0.33\textwidth}
        \includegraphics[width=\textwidth]{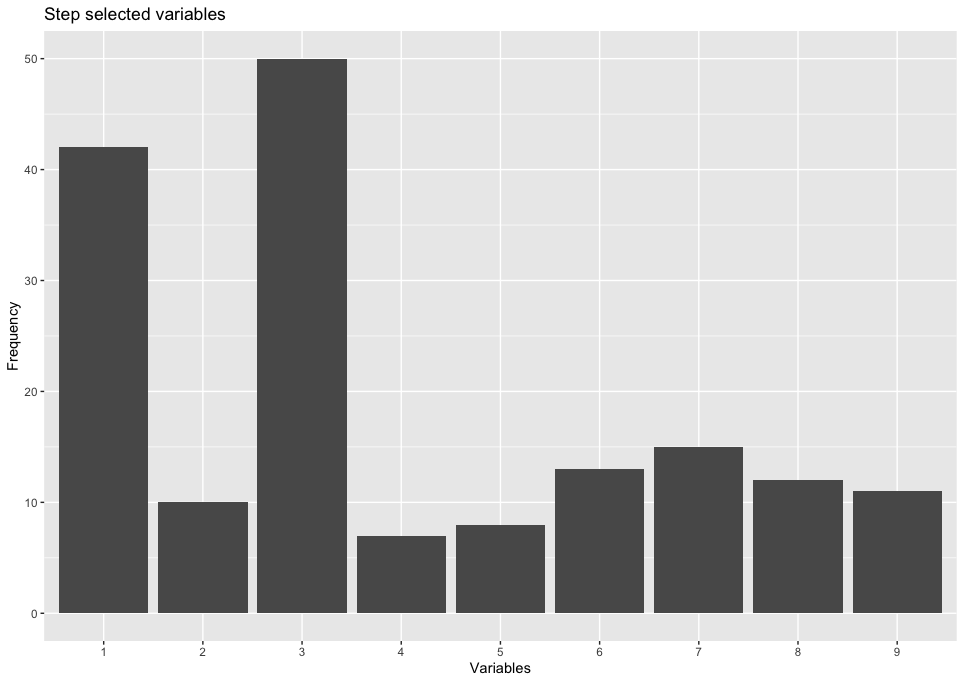}
        \caption{step}
        \label{fig:his_step_sim_re}
    \end{subfigure}
    \caption{comparison between kNNvs and stepwise logistic regression and the frequency of selected variables}
      \label{fig:his_com_sim_regression}
\end{figure}

Figure \ref{fig:his_com_sim_regression} shows the results of regression simulation study. Figure \ref{fig:compare with step sim for regression} is the 50 runs' MSE boxplots. This figure shows that the kNNvs has a better performance based on the median but a slightly larger range. The rest $2$ figures in Figure \ref{fig:his_com_sim_regression} show the frequency of selected variables. Both methods include some noise variables, but they both choose the most important variables $\vx_1$ and $\vx_3$. The other important variables are $\vx_7$ and $\vx_9$ in the simulation data set. kNNvs selects $\vx_9$ as the next important variable, whereas the step regression method picked $\vx_7$ more frequently than $\vx_9$ and the rest. So in this comparison, we could conclude that both methods are good, but kNNvs has the lower MSE. 

\subsubsection{Running time compare}
In this section, we simulate data with 500 observations. However, the p varies as shown in the table \ref{table:timetable}, all the variables are generated from multi-normal distributions, and the responses variable is generated from a Bernoulli distribution (according to equation \ref{eq:binarydata}), and we time the running time to see the efficiency of the $k$ nearest neighbors with grid search variable selection. Table \ref{table:timetable} shows that when p increases from 10 to 30, the running time increases from 3 to 33 secs. For p=30, which is a relatively large input space, kNNvs still can be run in around 30 secs which is an excellent machine. For the regression running time test, we downsize the sample size to 200 observations and compare the range of p from 10 to 30. The regression results are shown in table \ref{table:timetable-regression}; it shows that kNNvs could finish running in 1.2 seconds for $p=10$, even when the dimension goes up to 30, kNNvs still can be done almost in 10 seconds, which is a very efficient machine.    

\begin{table}[!ht]
  \caption{Classification Time table}
 \centering 
\begin{tabular}{rlrrrr}
  \hline\noalign{\smallskip}
  SN & Dataset  & $n$ & $p$  & Running Time\\ 
 \noalign{\smallskip}\hline\noalign{\smallskip}
1 & Data-1           &  500  &   10 &   3.23 secs\\
2 & Data-2           &  500  &   15 &   6.80 secs \\
3 & Data-3           &  500  &   20 &   13.89 secs\\
4 & Data-4           &  500  &   25 &   21.86 secs\\
5 & Data-5           &  500  &   30 &   33.08 secs\\
   \hline\noalign{\smallskip}
\end{tabular}
\label{table:timetable}
\end{table}

\begin{table}[!ht]
  \caption{Regression Time table}
 \centering 
\begin{tabular}{rlrrrr}
  \hline\noalign{\smallskip}
  SN & Dataset  & $n$ & $p$  & Running Time\\ 
 \noalign{\smallskip}\hline\noalign{\smallskip}
1 & Data-1           &  200  &   10 &   1.20 secs\\
2 & Data-2           &  200  &   15 &   1.74 secs \\
3 & Data-3           &  200  &   20 &   3.13 secs\\
4 & Data-4           &  200  &   25 &   5.24 secs\\
5 & Data-5           &  200  &   30 &   10.24 secs\\
   \hline\noalign{\smallskip}
\end{tabular}
\label{table:timetable-regression}
\end{table}

\subsection{Application on Real Data}

In this section, we apply the kNNvs model to real-world data. Table \ref{table:datasets} shows the data sets used in this section, including one regression and four classification data. The input space goes from 9 to 61, and the sample size is between 208 to 4177. Three evaluations are made in this section. The first is a predictive performance comparison between kNNvs and other machines;  The other two are to compare kNNvs on the predictive performance and selected variables for classification and regression cases. The data sets are provided by \citep{pei2021some}.
\begin{table}[!ht]
  \caption{Datasets table}
\centering  
\begin{tabular}{rlrrrr}
  \hline\noalign{\smallskip}
  SN & Dataset  & $n$ & $p$ &Data Type & $\kappa=n/p$ \\ 
 \noalign{\smallskip}\hline\noalign{\smallskip}
1 &  Asthmatic            &   405 &  11 &  Classification  & 36.82 \\
2 &  Congressional voting &   435 &  17 &  Classification & 25.59\\
3 &  Sonar                &   208 &  61 &  Classification &   3.41\\
4 &  Indian Liver Patient  &   538 &  11 &  Classification &    53\\
5 &  Abalone.              &   4177 &  9 &  Regression &    464\\
   \hline\noalign{\smallskip}
\end{tabular}
\label{table:datasets}
\end{table}

First, we compare kNNvs with some vary classic machines, namely support vector machine with the polynomial kernel, Gaussian kernel, and linear kernel, $k$ nearest neighbors model, and Gaussian process on the classification data sets. We randomly divide each data set into training data and test data for the experiment set-up, which is 70\% and 30\%. For the kernel and Gaussian-process models, we use cross-validation method to choose the best hyperparameters; for kNN and kNNvs, we use cross-validation to select the best k. Run the whole experiment 50 times, and compare the accuracy. 

\begin{figure}[!ht]
    \centering
    \begin{subfigure}[b]{0.49\textwidth}
        \includegraphics[width=\textwidth]{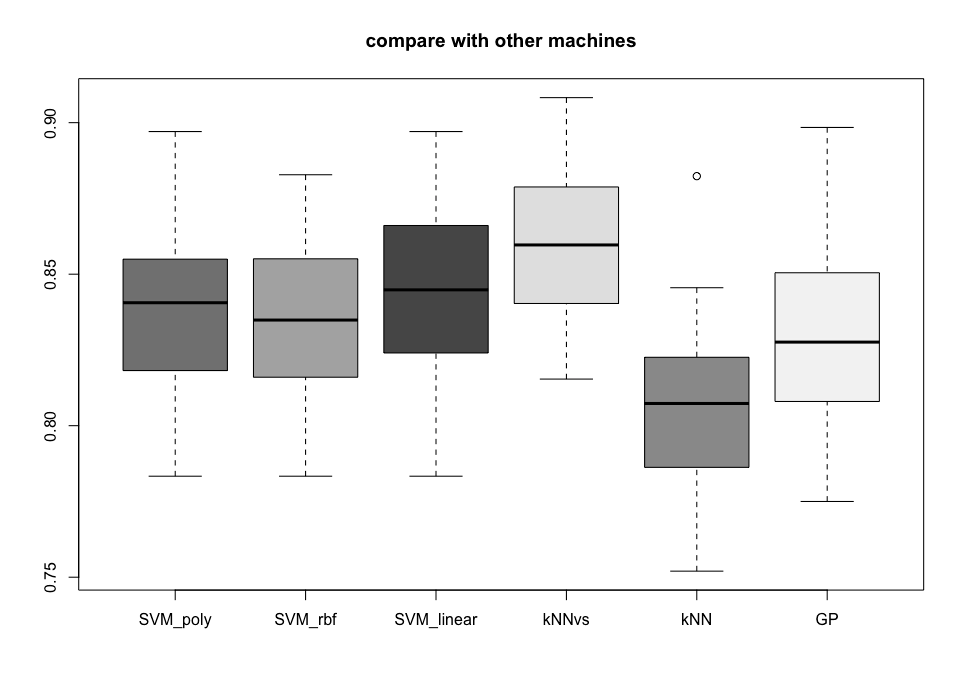}
        \caption{Asthmatic}
        \label{fig:Asthmatic}
    \end{subfigure}
        \begin{subfigure}[b]{0.49\textwidth}
        \includegraphics[width=\textwidth]{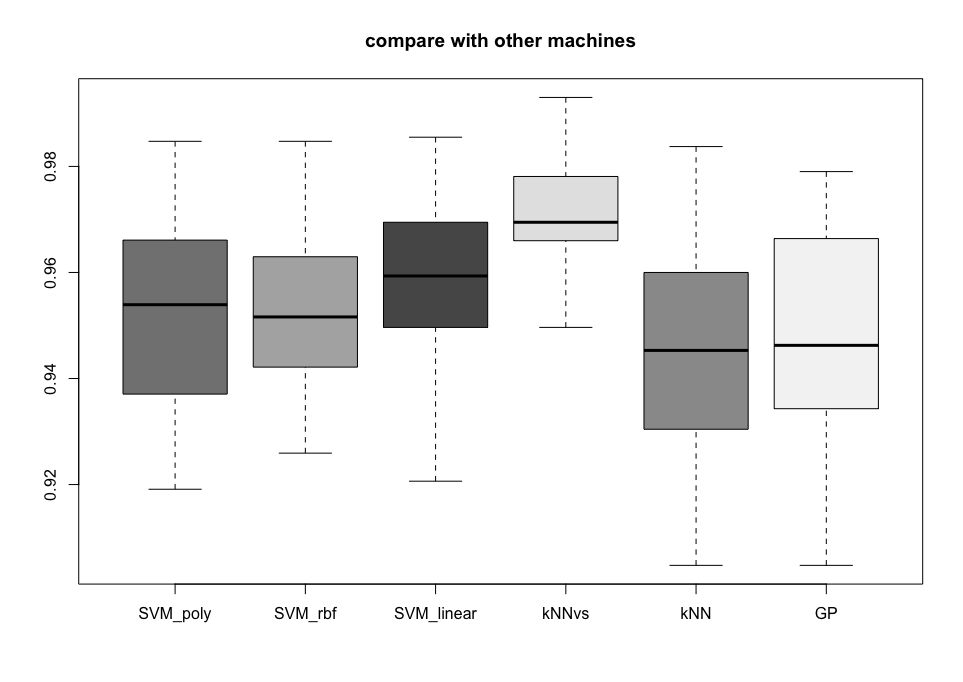}
        \caption{Voting}
        \label{fig:voting}
    \end{subfigure}
    \begin{subfigure}[b]{0.49\textwidth}
        \includegraphics[width=\textwidth]{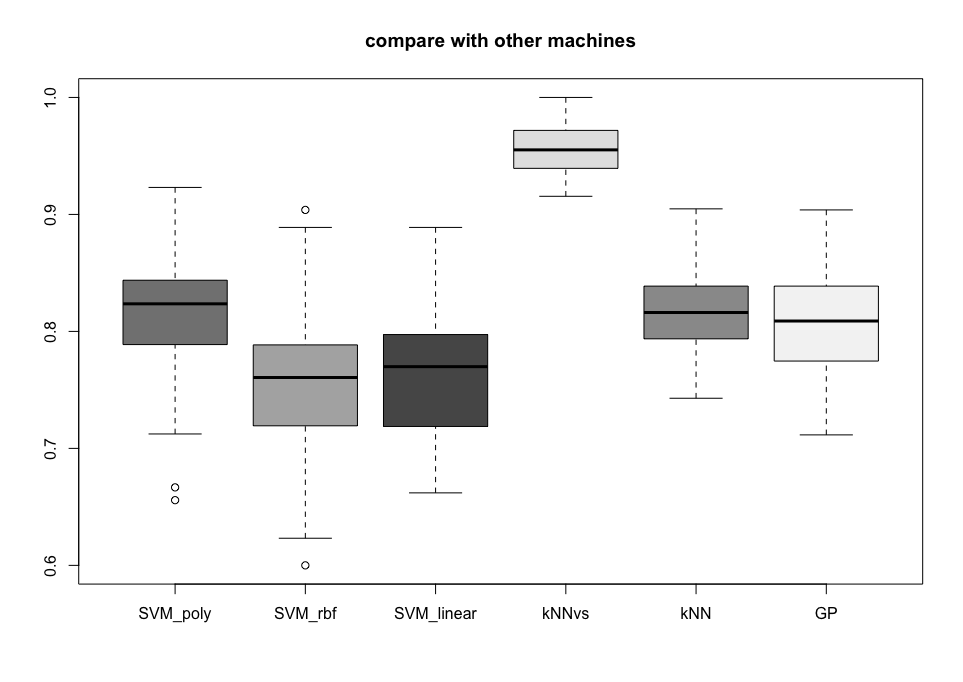}
        \caption{sonar}
        \label{fig:sonar}
    \end{subfigure}
    \begin{subfigure}[b]{0.49\textwidth}
        \includegraphics[width=\textwidth]{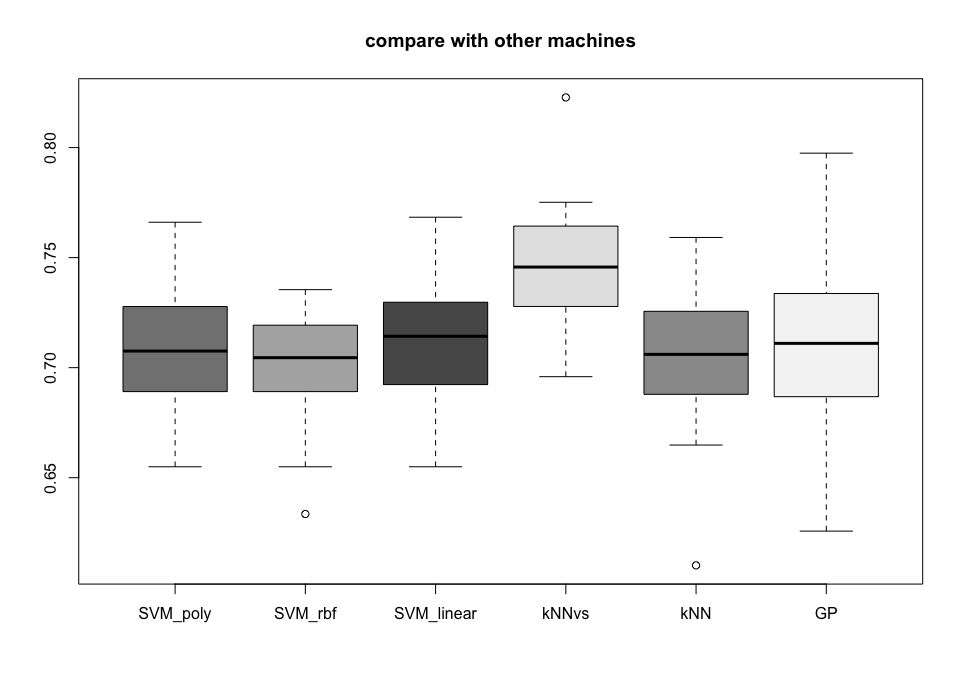}
        \caption{Indian Liver Patient}
        \label{fig:voting}
    \end{subfigure}

    \caption{comparison between kNNvs and other leaning machines}
      \label{fig:compare with other machines}
\end{figure}

Figure \ref{fig:compare with other machines} shows the box plot of the comparison results; the kNNvs is a lot better than kNN, and also, it works better than all the other powerful machines. Especially in Figure \ref{fig:voting}, kNNvs is doing a lot better than other machines and has the smallest range and better accuracy. One data in this experiment is so-called sonar data, with the most significant input space and smallest sample size, $\kappa$ = 3.41. Furthermore, among all the machines, kNNvs takes a small amount of time for tuning. 

The second experiment compares kNNvs with the stepwise variable selection method. In this experiment, we use the "Asthmatic dataset," and the data is divided into training and test data sets, which are 70\% and 30\%. Since this is a binary classification problem, the other machine using it is logistic regression which steps variable selection. This experiment runs 100 times.

\begin{figure}[!ht]
    \centering
    \begin{subfigure}[b]{0.32\textwidth}
        \includegraphics[width=\textwidth]{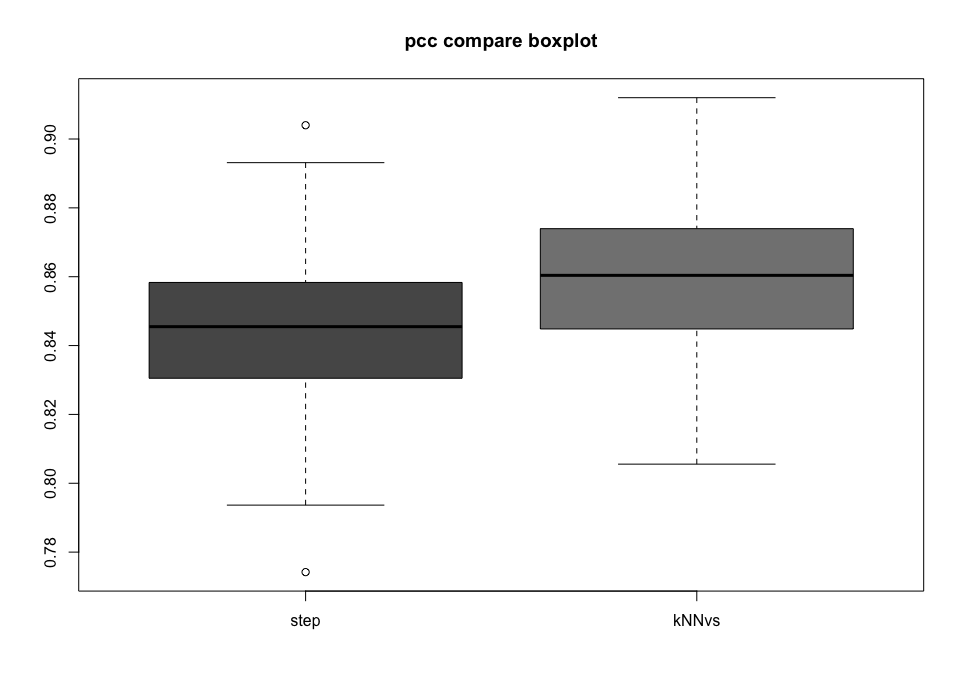}
        \caption{boxplot}
        \label{fig:compare with step}
    \end{subfigure}
    \begin{subfigure}[b]{0.33\textwidth}
        \includegraphics[width=\textwidth]{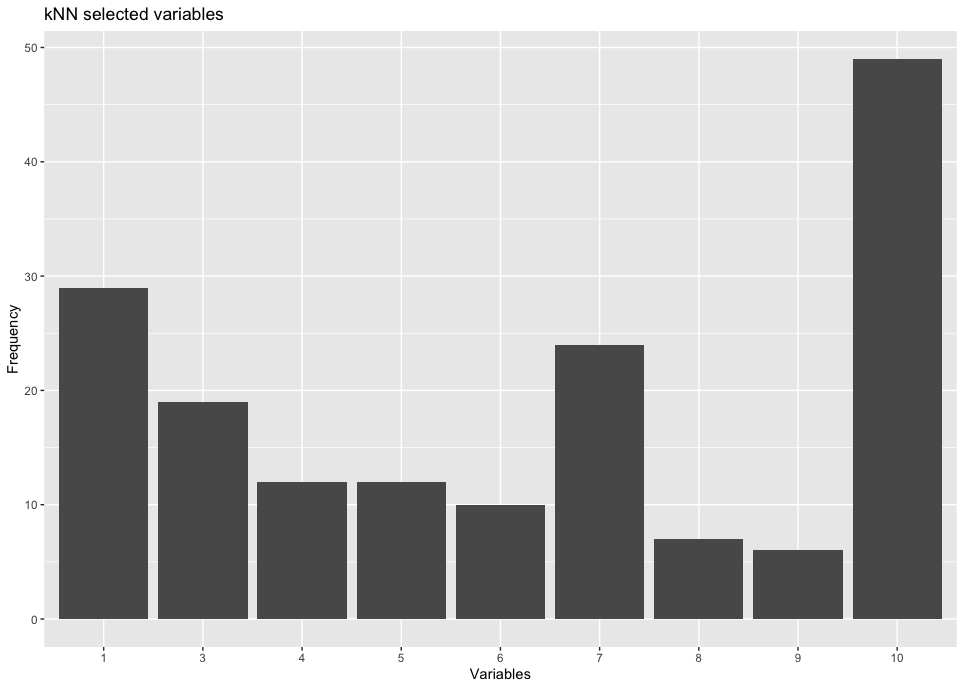}
        \caption{kNNvs}
        \label{fig:his_knnvs}
    \end{subfigure}
    \begin{subfigure}[b]{0.33\textwidth}
        \includegraphics[width=\textwidth]{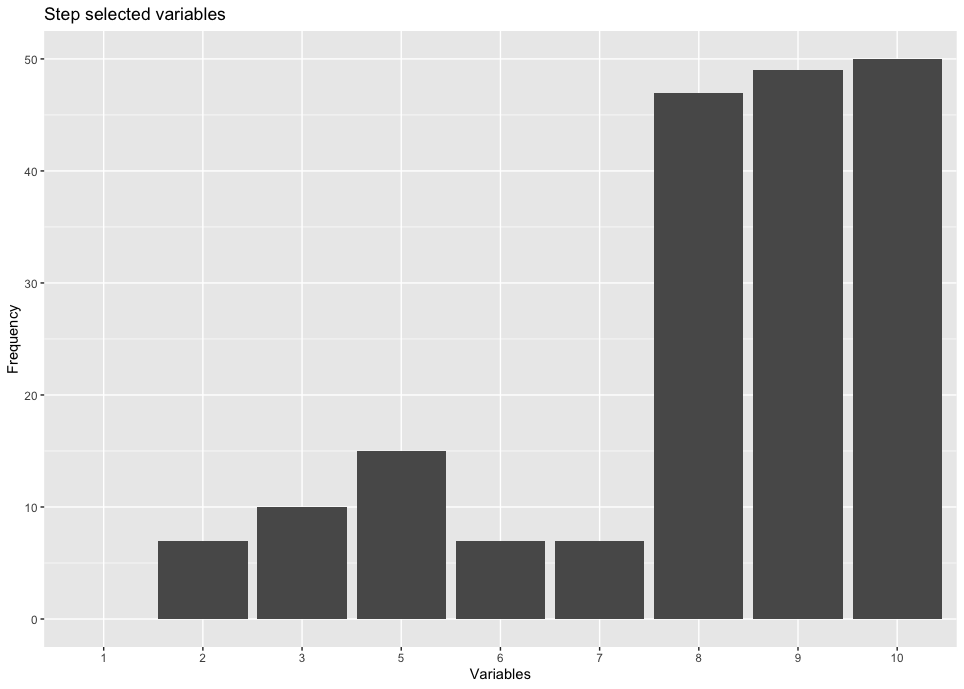}
        \caption{step }
        \label{fig:his_step}
    \end{subfigure}

    \caption{comparison between kNNvs and other step wise variable selection and frequency of selected variables}
      \label{fig:his_com}
\end{figure}

The first comparison is on the accuracy. Figure \ref{fig:compare with step} clearly shows kNNvs is better than step logistic regression on the predictive performance. The next evaluation is on the selected variables. Based on figure \ref{fig:his_com}, we could conclude that both methods agree on the essential variable, which is the 10th variable. However, The models selected the different second important variables. "step" chooses the 8th and 9th as the second important ones, whereas kNNvs chooses the 1st and 2nd as the second important ones. 


\begin{figure}[!ht]
    \centering
    \begin{subfigure}[b]{0.32\textwidth}
        \includegraphics[width=\textwidth]{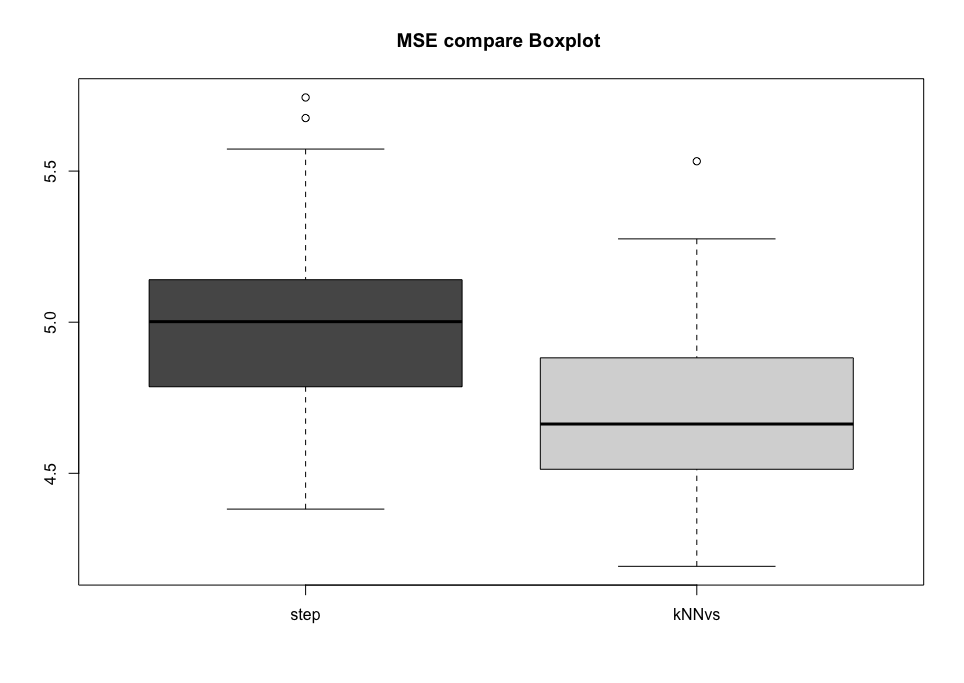}
        \caption{boxplot}
        \label{fig:regression compare with step}
    \end{subfigure}
    \begin{subfigure}[b]{0.33\textwidth}
        \includegraphics[width=\textwidth]{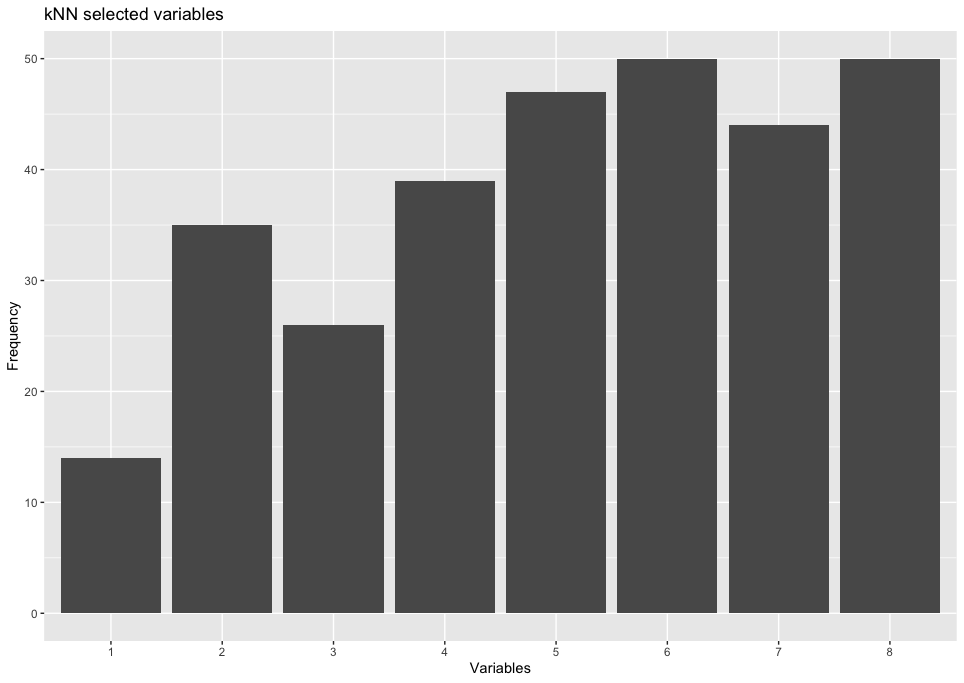}
        \caption{kNNvs}
        \label{fig:regression his_knnvs}
    \end{subfigure}
    \begin{subfigure}[b]{0.33\textwidth}
        \includegraphics[width=\textwidth]{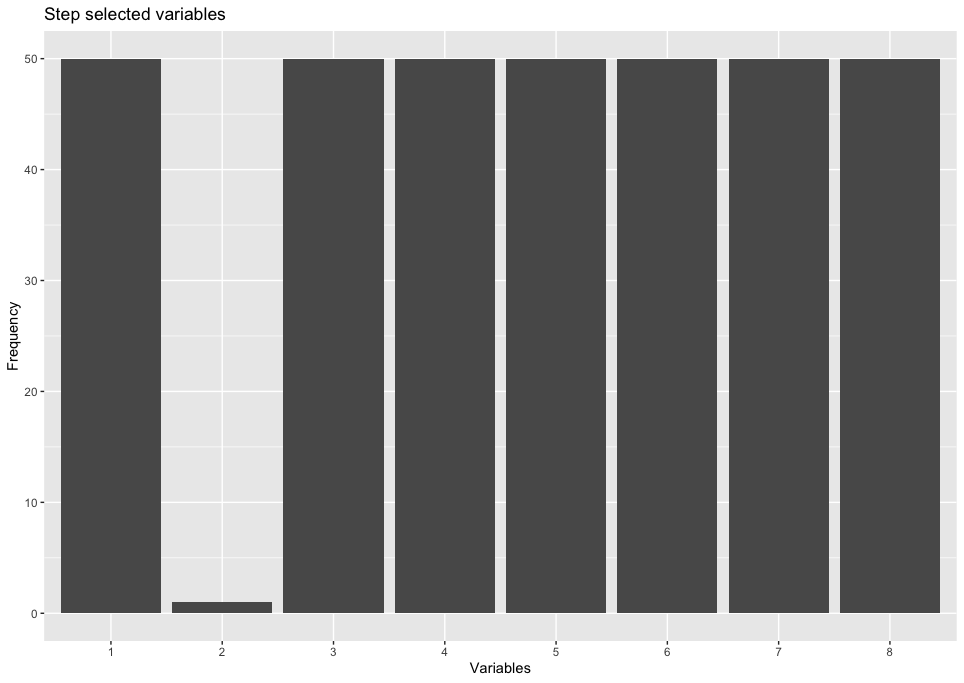}
        \caption{step }
        \label{fig:regression his_step}
    \end{subfigure}

    \caption{comparison between kNNvs and other step wise variable selection and frequency of selected variables on regression data}
      \label{fig:regression his_com}
\end{figure}

The last experiment is the comparison between kNNvs and step multi-linear regression. This experiment uses the regression data set Abalone dataset, which is large-size and high dimensional data. For the experimental set-up, similar to the others, we use 70\% as training and 30\% as test data, run the experiment 50 times, and compare the selected variables' frequency and the mean squared error(MSE). The results are shown in Figure \ref{fig:regression his_com}. Figure \ref{fig:regression compare with step} clearly shows kNNvs has a better performance than step regression. It has smaller MSE and fewer outliers. When coming to the selected variables, the step regression plot shows all the variables are equally important among 50 experiments but variable 2. However, the kNNvs frequency plots show that the last five variables are more important than the first three. So in this experiment, kNNvs has the better prediction performance. 

\section{Conclusion}
This paper proposes a new variable selection method for $k$ nearest neighbors, which combines grid search and forward variable selection. By using the real-world data and simulated data to prove our method has an improvement for original kNN, and also the prediction performance is better even than some mighty kernel machines. We made a particular simulation case on testing the running time, the KNNvs method took relatively less time to run some high-dimensional data. $k$ nearest neighbors with grid search variable selection can efficiently select the useful variables and provide a good prediction performance in both classification and regression cases.

\bibliographystyle{jds}
\bibliography{JDSbib}

\appendix
\section{kNNvs examples R code}
\begin{verbatim}
   library(kNNvs)
   data(iris3)
   train_x <- rbind(iris3[1:25,,1], iris3[1:25,,2], iris3[1:25,,3])
   test_x <- rbind(iris3[26:50,,1], iris3[26:50,,2], iris3[26:50,,3])
   cl_train<- cl_test<- factor(c(rep("s",25), rep("c",25), rep("v",25)))
   k<- 5
   # cl_test is not null
   mymodel<-kNNvs(train_x,test_x,cl_train,cl_test,k,model="classifiation")
   mymodel
   # cl_test is null
   mymodel<-kNNvs(train_x,test_x,cl_train,cl_test=NULL,k,model="classifiation")
   mymodel
\end{verbatim}

\section{A comparison about high correlated data}

The data are similar to the regression and classification simulation study set up but with different correlation between the input data. The results are similar. The kNNvs has a better performance.

\begin{figure}[!ht]
    \centering
        \begin{subfigure}[b]{0.32\textwidth}
        \includegraphics[width=\textwidth]{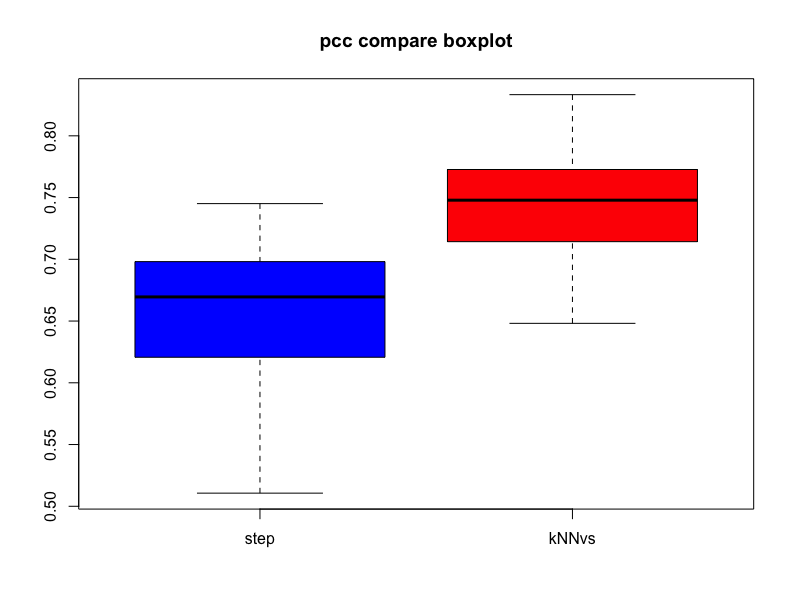}
        \caption{the boxplot }

    \end{subfigure}
        \begin{subfigure}[b]{0.33\textwidth}
        \includegraphics[width=\textwidth]{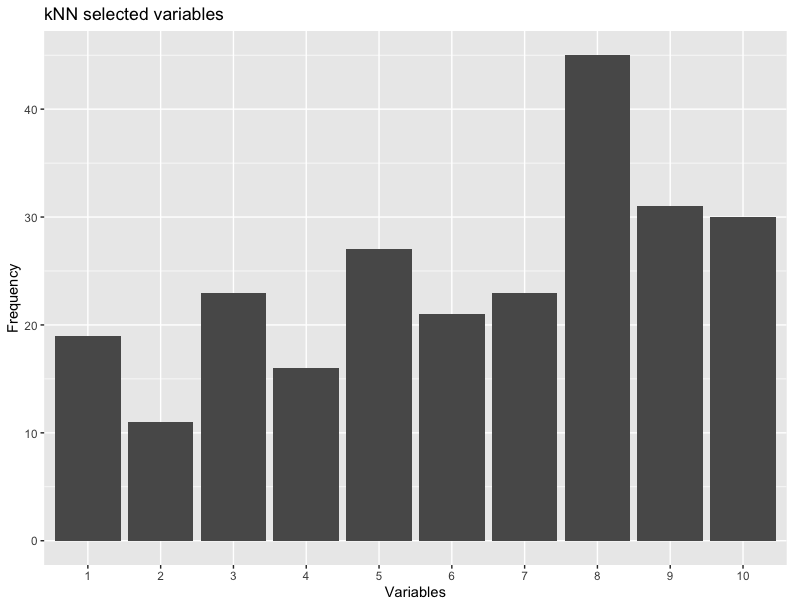}
        \caption{kNNvs }

    \end{subfigure}
    \begin{subfigure}[b]{0.33\textwidth}
        \includegraphics[width=\textwidth]{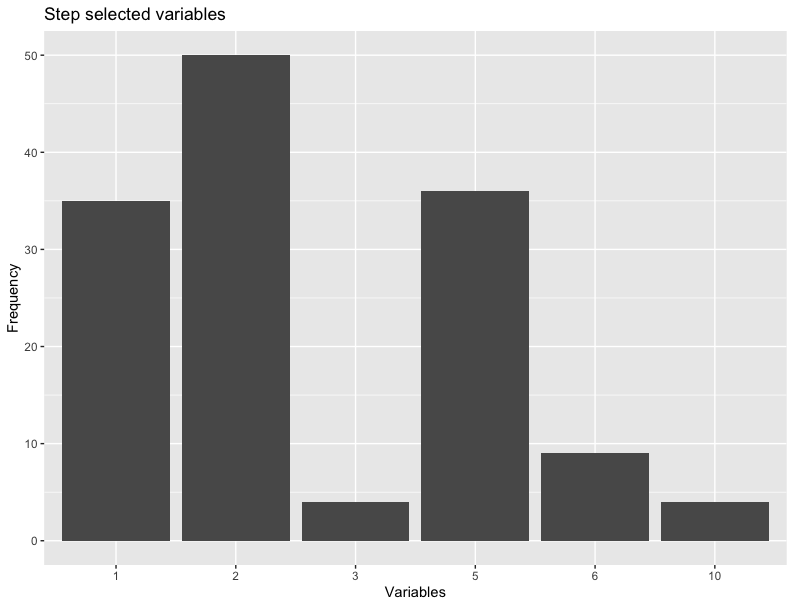}
        \caption{step}

    \end{subfigure}
    \caption{comparison between kNNvs and stepwise logistic regression and the frequency of selected variables,the correlation is set to 0.5}

\end{figure}

\begin{figure}[!ht]
    \centering
        \begin{subfigure}[b]{0.32\textwidth}
        \includegraphics[width=\textwidth]{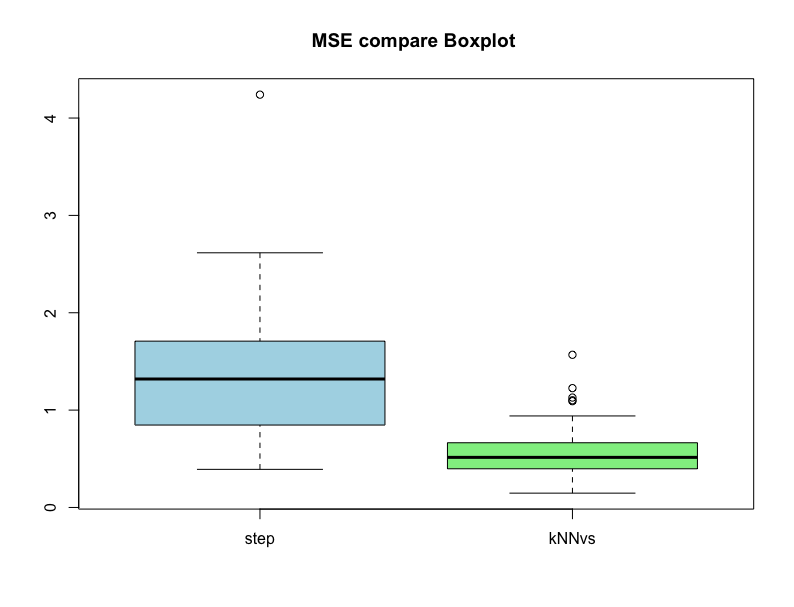}
        \caption{the boxplot }

    \end{subfigure}
        \begin{subfigure}[b]{0.33\textwidth}
        \includegraphics[width=\textwidth]{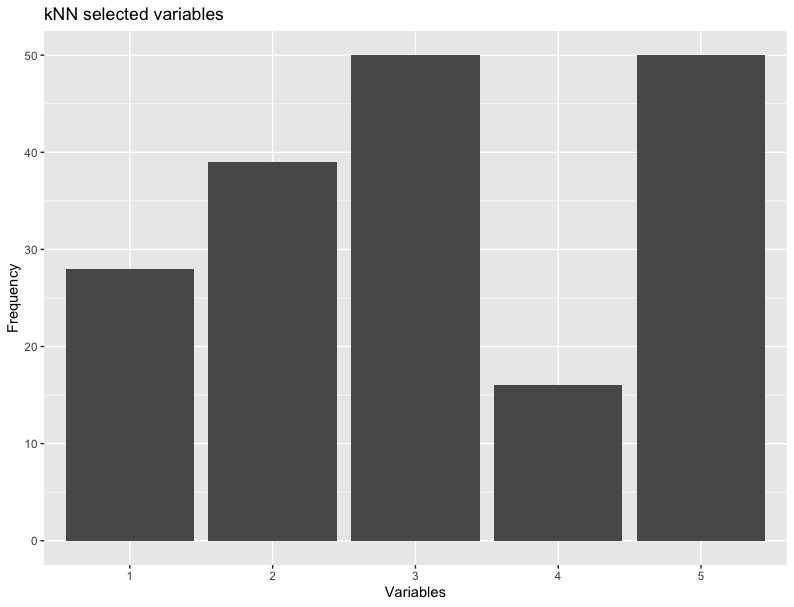}
        \caption{kNNvs }

    \end{subfigure}
    \begin{subfigure}[b]{0.33\textwidth}
        \includegraphics[width=\textwidth]{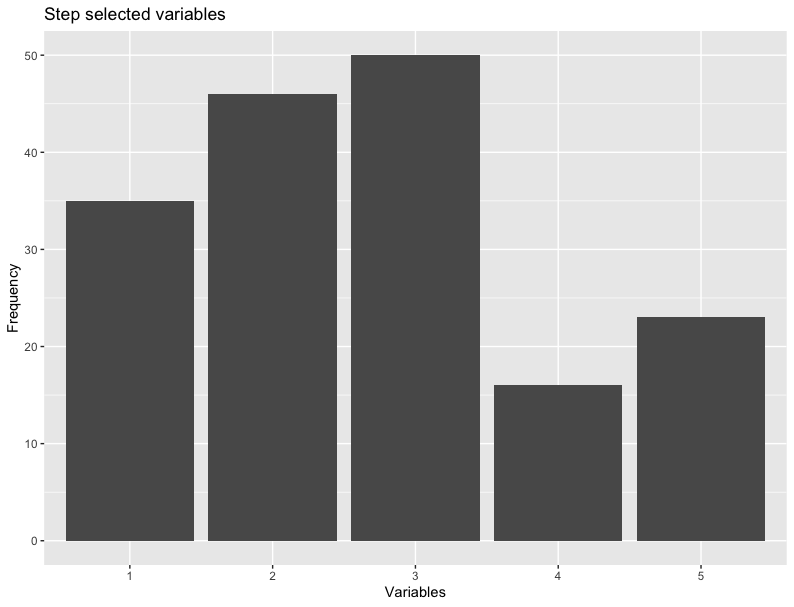}
        \caption{step}

    \end{subfigure}
    \caption{Regression results comparison between kNNvs and stepwise logistic regression and the frequency of selected variables,the correlation is set to 0.4}

\end{figure}

\end{document}